\documentclass{iaesarticle}

\usepackage{amsmath, amsfonts, amssymb, float, fancyhdr}
\usepackage[figuresright]{rotating}
\usepackage{authblk, graphicx, indentfirst, lastpage, lipsum}
\usepackage{pifont}

\makeatletter
\renewcommand{\@biblabel}[1]{[#1]\hfill}
\renewcommand\AB@authnote[1]{\textsuperscript{\normalfont\bfseries#1}}
\makeatother
\usepackage[font=normalsize]{subfig, caption}
\usepackage{epstopdf}
\usepackage[left=3cm, right=2.5cm, top=2.5cm, bottom=2.5cm, includehead, includefoot]{geometry}
\usepackage[justification=centering]{caption}
\usepackage{titlesec}
\usepackage[table]{xcolor}
\titleformat{\section}
{\normalfont\normalsize\bfseries\uppercase}{\thesection}{1.7em}{}
\titleformat{\subsection}
{\normalfont\normalsize\bfseries}{\thesubsection}{.95em}{}
\titleformat{\subsubsection}
{\bfseries}{\thesubsubsection}{.2em}{}
\titlespacing*{\section}{0cm}{0.55cm}{0cm}
\titlespacing*{\subsection}{0cm}{0.3cm}{0cm}
\usepackage{enumitem}
\usepackage[numbers,sort&compress]{natbib}
\usepackage{ragged2e}
\usepackage{hyperref,graphicx}
\hypersetup{hidelinks}
\usepackage{multirow}
\usepackage{fleqn}
\usepackage{tabularx}

\usepackage{array}
\newcolumntype{L}[1]{>{\raggedright\arraybackslash}p{#1}}
\newcolumntype{Y}{>{\raggedright\arraybackslash}X}
\setlist{nosep,leftmargin=1.5em,topsep=2pt,itemsep=1pt}

\makeatletter
\def\@CopyrightLine[#1]#2#3{\@copyrightlinetrue%
                       \gdef\@copyrightprefix{#1}%
                       \gdef\@copyrightyear{#2}%
                       \gdef\@copyrighttext{#3}%
                       \gdef\@CopyrightLine{\par\vskip0pt%
                       \noindent\ifx\@copyrightprefix\@empty\relax%
                       \else\@copyrightprefix~\fi%
                       ~\@copyrightyear~%
                       \@copyrighttext~}}
\makeatother

\CopyrightLine[]{}{\textit{This is an open access article under the \textcolor{blue}{\underline{CC BY-SA}} license.}
\vspace{.5em}}

\author[1]{\bfseries Nour-Eddine Moumni}
\author[1,2]{\bfseries Rachid Alaoui}
\author[3]{\bfseries Driss Kiouach}

\affil[1]{Laboratory of Research in Computer Science and Telecommunications (LRIT), Mohammed V University, Rabat, Morocco}
\affil[2]{Higher School of Technology of Sal\'e, Mohammed V University, Rabat, Morocco}
\affil[3]{Laboratory of Probability, Statistics and Intelligent Systems (LPAIS), Sidi Mohamed Ben Abdellah University, Fez, Morocco}

\title{Silent Reward and Environment Defects in EV Charging Simulation Benchmarks: A Diagnostic Checklist}
\shorttitle{Silent Reward and Environment Defects in EV Charging Benchmarks (Moumni)}

\begin{document}
\setcounter{page}{1}

\setlength{\parindent}{1.27cm}

\pagestyle{fancy}
\fancyhfoffset{0cm}

\journalname{Bulletin of Electrical Engineering and Informatics}
\journalshortname{Bulletin of Electr Eng \& Inf}
\journalhomepage{http://beei.org}
\vol{}
\no{}
\months{}
\years{}
\issn{}
\DOI{}
\pagefirst{}
\pagelast{}
\id{}

% arXiv preprint: the journal masthead below uses the volume/issue/page/DOI
% fields as they will eventually be assigned by BEEI at publication; since
% those are not yet known, the running header/footer is replaced here with
% a plain statement instead of leaving the official IAES template's
% unfilled placeholder fields visible.
\makeatletter
\fancypagestyle{firstpage}{%
\fancyhf{}
\lhead{\textbf{\@journalname} \\ Preprint (accepted manuscript); BEEI volume/issue not yet assigned.}
\rhead{\thepage}
\lfoot{\normalsize\emph{\textbf{Journal homepage:} \@journalhomepage}}
\rfoot{}
\renewcommand{\headrulewidth}{0.7pt}
\renewcommand{\footrulewidth}{0.7pt}
}
\fancypagestyle{followingpage}{%
\fancyhf{}
\fancyhead[LO]{\@journalshortname}
\fancyhead[LE]{\thepage}
\fancyhead[RO]{\thepage}
\fancyfoot[LE]{\@journalshortname --- preprint, accepted manuscript}
\fancyfoot[RO]{\emph{\@shorttitle}}
\renewcommand{\headrulewidth}{0.7pt}
\renewcommand{\footrulewidth}{0.7pt}
}
\makeatother
\pagestyle{followingpage}

\maketitle

\hrule
\vspace{.1em}
\hrule
\vspace{.2em}
\noindent
\parbox[t][][s]{0.315\textwidth}{%
\textbf{Article Info}
\vspace{.2em}
\hrule
\vspace{.2em}
\begin{history}
\vspace{.2em}

Received --, 2026

Revised --, 2026

Accepted --, 2026

\vspace{.3em}
\end{history}
\vspace{.2em}
\hrule
\vspace{.2em}
\begin{keyword}
\vspace{.2em}
Reinforcement learning \sep Reproducibility \sep Simulation benchmarks \sep Electric vehicle charging \sep SUMO \sep Reward shaping
\vspace{.2em}
\end{keyword}
\vspace{\fill}
}
\parbox{0.020\textwidth}{\hspace{0.5em}}
\parbox[t][][s]{0.65\textwidth}{%
\begin{abstract}
\vspace{.3em}
Reinforcement-learning (RL) benchmarks built on microscopic traffic
simulators such as SUMO are increasingly used to compare agent
architectures for electric vehicle (EV) charging assignment. Unlike
crashes, defects in the reward function, the environment's stochastic
model, or the state encoding often fail silently: the simulation
completes, produces plausible numbers, and the comparison is biased. We
report five such defects, found by auditing a five-agent benchmark on
OpenStreetMap road networks, stated generically: a shared random generator that
desynchronises seed-paired comparisons, an ablation-specific reward clip
applied unconditionally, a normalisation mismatch that zeroes a reward
term, a per-timestep penalty that should fire once per event, and a
state dimensionality documented but never wired into the network. We
quantify three: a rejection counter inflated up to 125x across
agents/cities, a reward term bounded at about $10^{-4}$ of its intended
scale, and a 16.6\%/16.3\% greedy-action divergence from a
state-encoding mismatch. Re-analysing a historical ablation, no A1--A2
effect survives family-wise correction; a 56-run campaign confirms the
RNG defect's mechanism directly while showing its correction does not
change that conclusion. We provide an eight-item checklist as a low-cost
first-pass audit; in a later regeneration, three checks caught two new
defects while three further process/artefact defects fell outside its
boundary.
\end{abstract}
}
\parbox[l]{\textwidth}{%
\rule{0.275\textwidth}{0.5pt} \hspace{0.5cm} \hrulefill
\\
\emph{\textbf{Corresponding Author:}}
\vspace{.2em}\\
Nour-Eddine Moumni\\
Laboratory of Research in Computer Science and Telecommunications (LRIT), Mohammed V University\\
Rabat, Morocco\\
Email: noureddine\_moumni@um5.ac.ma
}
\vspace{.2em}
\hrule
\vspace{.1em}
\hrule

\section{Introduction}

In EV-charging assignment studies, software-integrity defects are not
merely a reporting concern. They can alter the simulated station
availability seen by each policy, the number of rejected or served
vehicles, queue waiting time, completion rate, and delivered energy, while
the traffic simulation still terminates normally. Consequently, an
apparently plausible algorithm ranking may reflect inconsistent
electrical-service and transportation conditions rather than a policy
difference. This paper therefore treats implementation integrity as a
prerequisite for interpreting EV-charging performance, not as a
post-processing reproducibility check. In smart-city EV-charging
deployments, such a biased comparison would misinform exactly the
decisions these KPIs are meant to support --- charging-station
coordination, demand-responsive scheduling, and energy-management
policy --- even though no grid-level effect is measured in this paper.

While potential-based reward shaping guarantees policy invariance under precise theoretical conditions \cite{b28}, poorly designed reward shaping is known to induce reward hacking and
unintended policies \cite{b3}, formalised more recently as a general
property of proxy reward functions \cite{b7}. We examine two concrete
mechanisms in the audited EV-charging pipeline --- silent normalisation
mismatches and unconditional clipping outside their intended ablation
scope --- without claiming that these mechanisms are absent from prior
implementations or reports. Prior EV-charging RL studies \cite{b4,b5} report single-seed
or small-sample results, motivating the multi-seed guidance on adequately
powered comparisons \cite{b6}, which this paper's protocol adopts as its
non-parametric, seed-paired analysis unit (Sec.~\ref{sec:statistical-analysis}) without
itself claiming adequate power. In a
related EV-charging setting, Sadeghianpourhamami et al. \cite{b29}
formulate model-free reinforcement learning for coordination across
multiple charging stations. Qian et al. \cite{b25} couple grid and
transportation data in a single DRL charging-navigation state vector ---
exactly the kind of interface where the unit-consistency and
dimensionality checks we propose (D3, D5) apply --- and, beyond RL, Tadi
et al. \cite{b26} forecast demand and schedule charging slots with
supervised models, aggregating station-level capacity over predicted
arrivals and states of charge, exactly the kind of layer where the
counting and unit-consistency checks proposed here (D3, D4) would apply;
in neither case do we claim that the implementation contains these
defects. Where the field has invested
in shared, openly documented artefacts, ACN-Data \cite{b24} made real
workplace charging sessions inspectable and reusable on the data side;
this paper's checklist targets an analogous inspectability on the
simulator/reward/state side.

Comparative RL studies increasingly rely on stochastic microscopic traffic
simulators such as SUMO \cite{b1} --- coupled to external controllers via
interfaces such as TraCI \cite{b27} --- to evaluate charging-assignment,
routing, and scheduling policies for electric vehicles, following
established SUMO-based RL benchmark suites for traffic signal control and
mixed-autonomy traffic \cite{b8,b9}. These pipelines combine several
independently-seeded sources of randomness --- vehicle departure times,
driver routes, environment failure models, and the agents' own
exploration policies --- glued together through subprocess orchestration
and file-based configuration. This complexity creates a class of defects
qualitatively different from ordinary software bugs: they do not crash
the simulation, do not produce invalid numerical output, and the
resulting KPI tables look entirely plausible while simply biasing the
comparison between agents, often in a direction that happens to support
(or undermine) the paper's own hypothesis. Sculley et al. \cite{b23}
described a closely analogous failure mode in production machine-learning
systems, where configuration drift, undeclared data dependencies and
entangled components accumulate as ``hidden technical debt'' that
degrades system behaviour without raising an error, and Breck et al.
\cite{b30} proposed testing rubrics to audit production readiness against
exactly this risk; the setting differs from a research benchmark's
comparative bias, but both show defects remaining operationally silent.
The problem is systemic, not anecdotal: Semmelrock et al. \cite{b33} map
reproducibility barriers across the method, code, data and experiment
layers, and Gundersen and Kjensmo \cite{b22}'s survey of 400 AAAI and
IJCAI papers found none documenting all variables needed to reproduce its
results (only 20--30\% reported per factor) --- precisely the
documentation gap in which the defects described below survive
undetected. We show that even a rigorous multi-seed,
Wilcoxon-validated protocol \cite{b10} is vulnerable to defects that
undermine the very seed-pairing the statistical test relies on.

Henderson et al. \cite{b2} documented the broader reproducibility crisis
in deep RL, showing that unreported implementation details can reverse
published conclusions. Engstrom et al. \cite{b16} independently showed,
for policy-gradient methods, that undocumented code-level choices ---
rather than the headline algorithmic contribution --- account for most
of the reported PPO/TRPO performance gap, a code-level attribution
problem structurally analogous to the silent defects reported here.
Agarwal et al. \cite{b17} and Andrychowicz et al. \cite{b18} further
argue that small-sample point estimates routinely used to rank RL
algorithms are themselves statistically unreliable, reinforcing the case
for the seed-paired, non-parametric protocol we adopt (\cite{b10},
Sec.~\ref{sec:statistical-analysis}). Our contribution is complementary
and more operational: rather than surveying reproducibility in general,
we document five specific, recurring defect patterns in the reward
function, environment model, and state encoding of RL-traffic-simulation
pipelines, each illustrated with a real, quantified instance from our
five-agent EV-charging benchmark. What is new is not every check taken
alone --- several restate ordinary hygiene --- but their instantiation at
interfaces where a comparison can be silently biased, and the measured
cost of omitting them, generic enough to be plausible in any SUMO-based
(or comparable) multi-agent RL benchmark pairing simulation stochasticity
with per-agent exploration noise.

\subsection{Research questions and audit objective}
\label{sec:research-questions}

We frame this study around two research questions: \textbf{RQ1:} Which
silent implementation defects can invalidate cross-agent or
cross-configuration comparisons in a SUMO-based EV-charging benchmark, and
through which mechanisms? \textbf{RQ2:} Which low-cost, machine-checkable
invariants can detect these defects before KPI tables are interpreted? We
do not ask, and this single-project audit cannot answer, how frequently
these defects occur across the broader literature, or how well the
resulting checklist transfers to an independent codebase; both remain
future work (Sec.~\ref{sec:limitations}).

The contributions of this paper are: (i) a taxonomy of five silent defect
patterns (D1--D5), derived from one audited pipeline, affecting reward
functions, environment stochasticity, and state encoding in
RL-traffic-simulation benchmarks; (ii) quantified
evidence of impact from a real benchmark, namely an at-least-125x inflation
of a rejection counter (D4) and a reward term neutralised to about
$10^{-4}$ of its intended scale (D3); (iii) a re-analysis of a controlled
ablation showing that a partial reward fix can yield non-robust
directional signals on secondary KPIs --- none of which survives
family-wise correction --- while leaving the primary KPI unchanged
everywhere; and (iv) an
eight-item, actionable audit checklist for researchers building similar
pipelines.

\subsection{Related work and positioning}
\label{sec:related-work}

Prior work relevant to this audit spans three strands.
\emph{EV-charging assignment, routing, and reward shaping}
(Sadeghianpourhamami et al.~\cite{b29}; Qian et al.~\cite{b25}; Tadi et
al.~\cite{b26}; ACN-Data~\cite{b24}) is developed above in the
Introduction; \emph{SUMO-based traffic-simulation benchmarks} and
\emph{reproducibility and statistical rigour in RL} supply the
remaining substrate and methodological backdrop, likewise introduced
above. All three strands are compared work-by-work in
Table~\ref{tab:related-work-comparison} below.

Each strand, however, risks studying its own object without auditing the
software interfaces where all three meet.
We quantify that gap over a
representative, not exhaustive, sample of ten works: those cited
above that describe their own artefacts or process --- an experiment, a
report, or a checklist programme --- in enough textual detail to be
checked against the five criteria. Table~\ref{tab:related-work-comparison}
below reports the full ten-work comparison directly, merging works that
share an identical RNG/reward/state combination into one row for
compactness (per-cell source citations for every entry are archived; see
Data Availability). Cited foundational/theory works,
external surveys~\cite{b22,b33}, and non-selected empirical
studies~\cite{b16,b18} are excluded without prejudice to their
protocols. Each cell is individually verified against the cited work's
full text and traceable either to a direct citation or to an explicit
negative search record. Among these ten
works, seed/uncertainty reporting is present in half (5/10),
environment--agent RNG parity in one (1/10, Patterson et
al.~\cite{b21}), reward-unit/ablation-integrity checks in three (3/10),
and state-schema checks in six (6/10); among the ten prior works, only
Patterson et al.\ report all three integrity checks jointly, and none does so within a traffic- or
EV-charging-simulation pipeline. Beyond the contributions already stated
(Sec.~\ref{sec:research-questions}), we claim no novelty in RL
reproducibility methodology in general, in the SUMO/TraCI substrate, or
in the specific EV-charging algorithms used; and beyond the ten works
audited individually here, we do not claim, and have not individually
verified, whether any specific study does or does not audit these
interfaces.

\begin{table}[!ht]
\centering
\fontsize{7.5pt}{9.5pt}\selectfont
\caption{Related-work comparison across all ten audited works, merging
works that share an identical RNG/reward/state combination into one row.
R = reported, N = not reported, U = unclear/not verified.}
\label{tab:related-work-comparison}
\begin{tabularx}{\textwidth}{p{2.15cm}p{2.0cm}p{1.7cm}p{0.9cm}p{1.9cm}p{1.9cm}X}
\hline
\textbf{Study} & \textbf{Setting} & \textbf{Method} & \textbf{Stat.\ rep.} & \textbf{Integrity checks (RNG / reward / state)} & \textbf{Pipeline scope} & \textbf{Gap vs.\ present study} \\
\hline
Patterson et al.~\cite{b21} & RL empirical design, gridworlds/MuJoCo & E-Sarsa, DQN, EQRC, SAC, DDPG & R & R / R / R & General RL (gridworlds/MuJoCo) & no traffic/EV interface \\
Ault \& Sharon; Vinitsky et al.~\cite{b8,b9} & Traffic-signal and mixed-autonomy traffic, SUMO/Flow & IDQN, IPPO, MPLight, FMA2C; ARS, ES, TRPO, PPO & R & N / R / R & SUMO traffic (non-EV) & no EV-charging arm \\
Qian et al.~\cite{b25} & EV charging navigation & DQN + DSCRM & N & N / U / R & EV-charging (non-SUMO) & reward check unclear, not reported in text \\
Tadi et al.~\cite{b26} & EV smart charging, energy harvesting & Random Forest, K-means, LSTM & N & N / N / U & EV-charging (non-SUMO) & RNG/reward not reported in text \\
Lee et al.\ (ACN-Data); Sadeghianpourhamami et al.~\cite{b24,b29} & Workplace EV charging prediction; EV charging coordination & Gaussian mixture models; fitted Q-iteration & N & N / N / R & EV-charging (non-SUMO) & RNG/reward not reported in text \\
Agarwal et al.; Pineau et al.~\cite{b17,b20} & RL methodology/evaluation; ML reproducibility policy & N/A (survey); N/A (policy report) & R / N & N / N / N & General RL/ML methodology & no integrity checks reported in text \\
Henderson et al.~\cite{b2} & RL methodology, reproducibility crisis & TRPO, DDPG, PPO, ACKTR & R & N / U / N & General RL & RNG/state not reported in text \\
\textbf{Present study} & EV charging, silent-defect audit, SUMO & DQN, DDQN, Tabular Q-L, Heuristic, Random & R & R / R / R & SUMO EV-charging pipeline & audits RNG/reward/state jointly \\
\hline
\end{tabularx}
\end{table}

\section{Method}

\subsection{Benchmark and EV-charging environment}

The benchmark simulates EV charging assignment with SUMO on real
OpenStreetMap road networks of two Moroccan cities, Rabat and Tangier,
under distinct traffic periods per city (morning, midday, evening peaks),
giving six (city, period) scenarios in total. Each scenario couples
vehicle demand, road-network stochasticity, and a fixed set of charging
stations to a policy under evaluation; the queue-wait time, completion
rate, delivered energy, and vehicles-served KPIs reported throughout this
paper (Sec.~\ref{sec:results}) are computed per episode from this
simulated environment. The audit uses a 13-dimensional implemented state,
four station actions, the declared reward terms, and a 40-episode training
plus 15-episode evaluation schedule; the complete parameter set (simulator version, network
geometry, optimiser/dependency detail) is Table S3,
referenced explicitly rather than restated here. Finer network geometry
(individual lane/junction coordinates, per-vehicle routes) lives in the
archived SUMO scenario configuration files, and the
campaign manifests separately record the full agent/scenario/seed grid.

\subsection{Agent selection and training protocol}

We audited five agents: a random baseline and an adaptive heuristic
baseline as non-learning controls, a tabular Q-learning agent
\cite{b11}, a Deep Q-Network (DQN) agent \cite{b12} with a
uniformly-sampled experience-replay buffer, and a Double DQN (DDQN) agent
\cite{b13} addressing DQN's overestimation bias. This diversity is
methodologically useful for the audit because the agents consume
different numbers of random draws and use different state and reward
representations, thereby exposing parity failures such as D1 and D3
(Sec.~\ref{sec:taxonomy-intro}). It is not intended as an exhaustive
sample of contemporary RL algorithms, and no claim of
algorithm-independence is inferred from these five agents alone. Seed
counts and the scenario/agent protocol are recorded in the archived
campaign manifests.

\subsection{Audit procedure and defect-confirmation rule}
\label{sec:audit-procedure}

The audit compared, for every agent, (i) the source of randomness driving
environment stochasticity against the source driving agent-internal
decisions, (ii) every ablation feature flag against every operation it was
expected to gate, (iii) every reward-function input against its
corresponding state-vector representation to check for normalisation
consistency, (iv) every counter and penalty tied to a discrete simulation
event against the control-flow condition governing it, and (v) the
state-vector dimensionality actually constructed at runtime against the
dimensionality documented in the project's companion papers. A defect was
confirmed only once all four of the following were established: the
expected invariant, the observed violation traced to a specific code
path, a minimal reproducing condition, and a fix locked in by a
regression test (Sec.~\ref{sec:results}, \ref{sec:prospective}). Each
finding was, where feasible, additionally quantified using production run
data recorded before the corresponding fix was applied. We additionally
assign each defect an evidence-strength tag on a three-level scale ---
deliberately not a severity or real-world-impact ranking, which this
single-project audit cannot establish: \emph{quantified} (a direct
KPI-level or counter-level effect is measured, e.g.\ D4), \emph{propagating}
(the defect corrupts a shared resource --- the RNG stream or the learning
signal --- so every downstream KPI is plausibly affected without an
isolated causal estimate, e.g.\ D1, D2, D3), and \emph{structural} (a
comparability or architecture mismatch is established without a measured
KPI effect, e.g.\ D5's documented-vs-implemented dimensionality). A
\emph{propagating} tag does not imply a larger real-world consequence
than a \emph{quantified} one; it reports that the causal chain to KPIs is
plausible but unmeasured, which is a statement about the evidence
available, not about impact. A severity scale proper --- ranking defects
by consequence rather than by evidence strength --- would need at least
two ingredients this single-project audit does not have: an established
KPI-causal magnitude for every defect (available only for D4 and, at the
decision level, D5) and a stated deployment or safety threshold past
which a given KPI shift becomes consequential for EV-charging operations,
neither of which this paper measures or assumes.

\subsection{Statistical analysis}
\label{sec:statistical-analysis}

The paired analysis unit throughout this paper is the random seed. Seven
seeds per configuration were the campaign's fixed compute budget, set
before either the ablation or the audit reported here; we do not claim
this budget was chosen for adequate power, and Sec.~\ref{sec:results}
discusses its consequence for the resolution of the test explicitly. The
declared A1--A2 comparison family (Sec.~\ref{sec:results}, Table S2) is
six scenarios $\times$ four KPIs $=24$ tests; the D5 secondary-endpoint
family (Sec.~\ref{sec:taxonomy-intro}) is four KPIs $\times$ two agents
$=8$ tests. Within each family, $p$-values are computed by exact Wilcoxon
signed-rank enumeration over the $2^{n}$ sign assignments for $n=7$
matched seed pairs -- a conditional, distribution-free test of the null
hypothesis that the sign of each pair's difference is equally likely,
which does not require the paired differences themselves to be normally
distributed, only exchangeable in sign under the null. Zero differences
are excluded from the ranked set before enumeration, following the
standard Wilcoxon convention (none occur in this paper's declared
families); where non-zero differences tie in absolute value, ranks are
averaged (mid-rank) before enumeration, which this paper's declared
24-comparison family requires for exactly one cell (vehicles-served,
Tangier midday, where two seeds' paired differences are exactly equal in
absolute value); every other cell, including a near-equal pair at
vehicles-served/Tangier morning that agrees only to 15 significant
figures, has strictly distinct, non-zero absolute paired differences, as
verified directly on the released seed-level data and applied
identically by the released recomputation script. Holm--Bonferroni correction is applied across the
full declared family, never mixed across families. Where reported, paired
mean differences receive a percentile bootstrap 95\% confidence interval
(10,000 resamples, fixed RNG seed), and Cohen's $d_z$ (mean paired
difference over its standard deviation) is reported separately alongside
it (Fig.~\ref{fig2}); the full 24-comparison Table S2 reports
direction and $p$-values only, without bootstrap CIs or $d_z$ for every
cell. We treat a
corrected non-significant result as insufficient evidence of an effect at
this seed resolution, not as evidence of no effect
(Sec.~\ref{sec:results}).

\subsection{Taxonomy of defects}
\label{sec:taxonomy-intro}

\begin{figure}[!t]
\centering
\includegraphics[width=\linewidth]{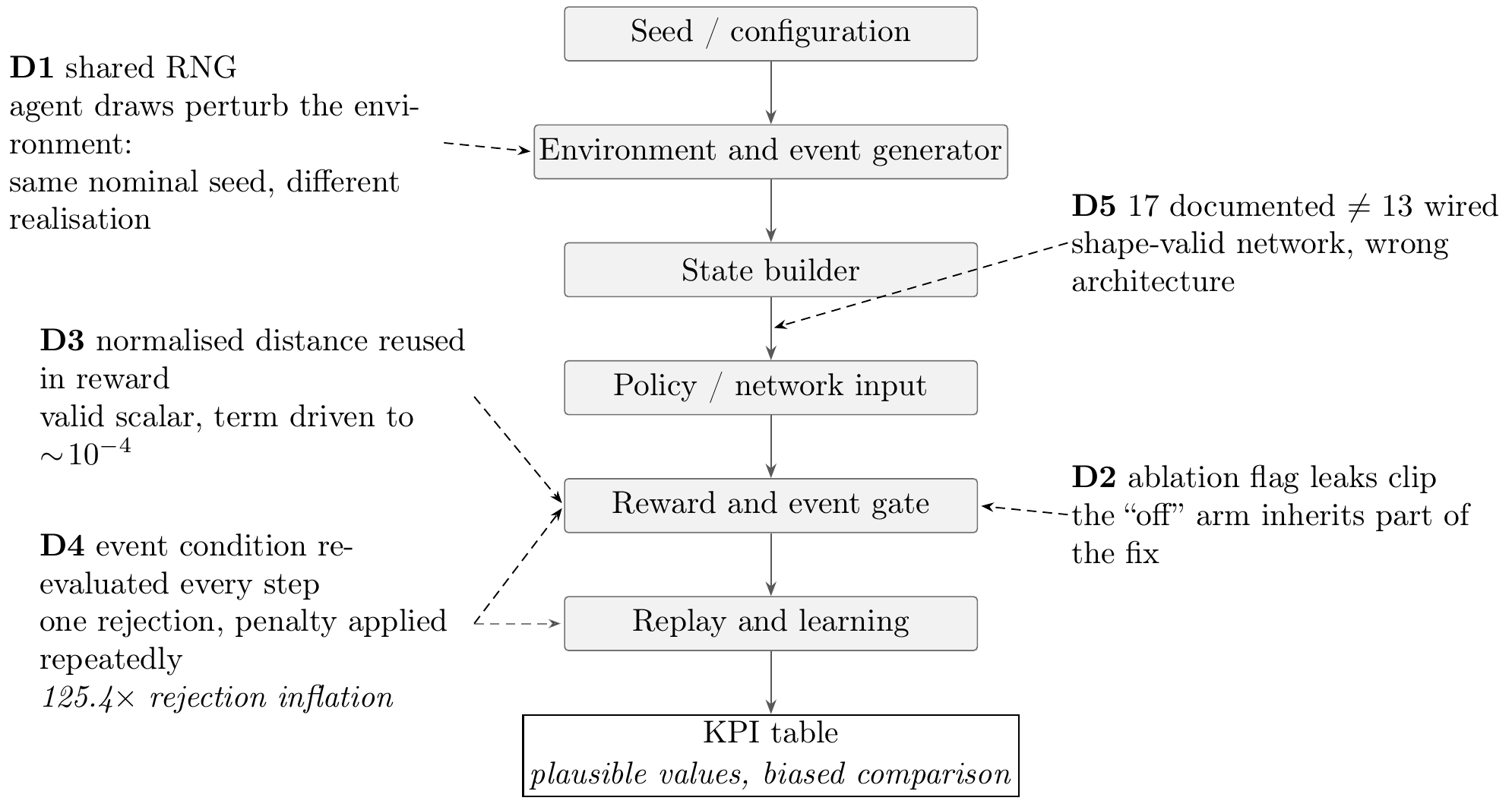}
\caption{RL-traffic-simulation pipeline from seed/configuration to KPI output,
showing silent-defect points D1--D5. Solid arrows indicate nominal flow;
dashed arrows indicate defect mechanisms. All paths yield plausible KPI
tables without exceptions.}
\label{fig1}
\end{figure}

Each defect is stated generically below, then illustrated with the
concrete instance found in our pipeline. Figure~\ref{fig1} locates each
defect at the pipeline stage it affects.

\textbf{D1 --- Shared RNG invalidates seed pairing (evidence:
propagating).} When one global generator is seeded per episode and shared
by environment stochasticity and agent randomness, agents that consume
different numbers of draws observe different environment realisations under
the same nominal seed. In our pipeline, the two-state station-failure
Markov chain shared the stream with exploration and replay sampling; DQN
draws up to 64 values per learning step versus one for the random-policy
baseline, so failure sequences diverged from the first learning step.

\textbf{D2 --- Reward clip outside its ablation condition (evidence:
propagating).} A feature-flagged reward fix can leak a clipping or rescaling
side effect into the nominally pre-fix arm. In our pipeline, the bound
$[-2,+1]$ was applied unconditionally in the DQN/DDQN caller immediately
after the shared reward function returned, while the rest of the fix was
flag-gated.

\textbf{D3 --- Normalisation mismatch silently zeroes a reward term
(evidence: propagating).} When
a state observation is normalised for neural-network input and the same
normalised value is later reused as an input to the reward function ---
which applies its own, independent normalisation --- the result is
double normalisation: a term intended to range over $[0,1]$ collapses to
a negligible magnitude, becoming numerically inert without raising any
error. In our pipeline, a route-distance feature already normalised to
$[0,1]$ for the state vector was passed unchanged into the reward
function, which itself divided by a detour-normalisation constant of the
same order of magnitude, scaling the resulting term by about $10^{-4}$
relative to its intended weight for the two neural-network agents
sharing this representation, while a tabular agent receiving the
distance in unnormalised metres retained a fully active detour term.
Concretely, the intended detour weight is $W_{\text{DETOUR}} = 1.0$ and
the reward-side normalisation constant is $\text{DETOUR\_MAX} =
10{,}000$ (config.py); the defect affects
specifically the charging-completion (terminal) transition, whose
detour term is drawn from a distance already normalised once by the
state-side \texttt{MAX\_DIST} (6{,}500~m Rabat, 2{,}000~m Tangier), then
divided by $\text{DETOUR\_MAX}$ a second time, bounding it at
$1/\text{DETOUR\_MAX} = 1\times10^{-4}$ regardless of which
city-specific constant produced the state-side value --- a strict upper
bound, not a distribution, since the archived evidence preserves the
code and its constants but not a transition-level reward log at that
granularity. By contrast, the periodic (every-10-step) reward computed
while a vehicle is still en route queries road distance directly in
metres, bypassing the state-normalised value entirely; the defect
therefore stays confined to the terminal transition. Four other reward
terms are active in the same configuration ---
$W_{\text{SUCCESS}}=5.0$, $W_{\text{QUEUE}}=5.0$ (quadratic),
$W_{\text{SERVICE}}=1.5$, $W_{\text{CONGESTION}}=2.0$ --- each three to
five orders of magnitude larger than the $10^{-4}$ bound on the detour
term, so on the terminal transition the latter's maximum possible
contribution is negligible beside them. No new simulation was needed to
verify this --- only re-reading the normalisation constants already
present in the archived pre-audit code. \emph{Fix:} see Table~\ref{tab:checklist}, check~3.

\textbf{D4 --- Per-timestep penalty instead of per-event penalty (evidence:
quantified).} If a persistent rejection or failure condition is updated
inside the per-step loop without a once-per-entity guard, one logical event
is counted and penalised repeatedly. In our pipeline, a vehicle unable to
secure a station was re-evaluated every 1-second step; each step incremented
the rejection counter and added a negative-reward transition to replay.

\textbf{D5 --- State dimensionality documented but never wired (evidence:
structural).} A
paper's Methods section and a project's code can drift apart silently
when a described input feature is defined in configuration, imported
into the relevant module, but never actually concatenated into the state
vector passed to the network, instantiating the formalism--implementation
gap framed by Castro~\cite{b31}. In our pipeline, an earlier documented
specification and the implemented network input diverged: the
specification described a 17-dimensional state vector including four
station-power features, while the corresponding neural-network
implementation used a 13-dimensional input, so the trained networks never
used the specified architecture.
D5's fix is implemented in the audited pipeline and locked by regression tests;
the corresponding implementation checks are summarized in
Table~\ref{tab:checklist}.

\textbf{D5 decision-level quantification.} To quantify D5's decision-level
impact, we generated a frozen decision corpus: a provenance-locked,
deterministic post-hoc replay of the control-arm DQN (seed 45,
rabat\_evening, 15 greedy evaluation episodes; not an extraction of a log
from the original campaign, since none was produced --- see Data
Availability). This yields a single shared sequence of 499 states and
action masks on which the matched 13D and 17D checkpoints from all 7
training seeds are scored; reported confidence intervals therefore capture
checkpoint-to-checkpoint variability conditional on this fixed decision
sequence, not variability across independent evaluation trajectories. The
greedy actions disagree at 16.6\% (DQN, 95\% CI [13.9\%, 19.7\%]) and
16.3\% (DDQN, 95\% CI [13.0\%, 18.8\%]), reported descriptively. None of
the 4 downstream KPIs (queue-wait time, completion rate, delivered energy,
vehicles served) shows a Holm-corrected significant difference across the
8-test family (smallest $p_{\text{Holm}} = 0.500$), though DQN completion
rate and delivered energy show the largest paired effects ($d_z \approx
1.0$, $p_{\text{raw}} \in [0.06, 0.08]$), not interpreted here as an
established causal effect.

D1 affects the comparability of the failure conditions each agent is
compared under --- and therefore, indirectly, every KPI that depends on
station availability --- without being isolated as a standalone causal
effect on any one of them in the historical archive; Sec.~\ref{sec:a7-validation}
tests this directly and finds no measurable change to the historical
ablation's conclusion --- a robustness finding for that one ablation,
not a general claim that D1 has no effect anywhere it could occur. D2 and D3 alter the learning
signal and can plausibly displace queue-wait time, completion rate,
delivered energy, or vehicles served, without their causal effect being
quantified here; because both bias the reward an agent optimises for
congestion and detour, the engineering decision most at risk is
charging-station assignment under load, not a single KPI in isolation.
D4 directly corrupts the rejection counter and the replay-buffer
transitions it feeds, and its magnitude is quantified in full
episode-level detail for one representative configuration and as a mean
across the four non-random agents and both cities (Sec.~\ref{sec:results}). D5
changes the DQN/DDQN state encoding, omitting four station-power
features a real power-aware charging-coordination decision would need;
its decision-level effect is quantified, but its downstream KPI
effect is not statistically distinguishable from zero at $n=7$ (D5
decision-level quantification paragraph above). We do not convert D1,
D2, D3, or D5's plausible mechanisms into established KPI-level effects
beyond what is directly measured, nor any grid-level or
station-utilisation-rate effect beyond the KPIs already reported.

\subsection{Consolidated defect-impact and detection table}
\label{sec:consolidated-table}

Table~\ref{tab:d1d5-mechanism} summarises, for each defect D1--D5, the
affected component, mechanism, expected versus observed behaviour,
measurable impact, and affected agents, answering the
mechanism-and-evidence framing requested in review. The corresponding
detection-and-remediation framing is answered by
Table~\ref{tab:checklist} below, whose \emph{Corrective action} column
gives the same fixes alongside the checklist that encodes them as
executable assertions, rather than restating them in a separate panel.

\begin{table}[!htbp]
\centering
\caption{Defect mechanism and evidence.}
\label{tab:d1d5-mechanism}
\fontsize{7.5pt}{9.5pt}\selectfont
\begin{tabularx}{\textwidth}{L{0.4cm}L{2.9cm}L{3.2cm}YL{2.8cm}}
\hline
\textbf{D} & \textbf{Component / mechanism} & \textbf{Expected vs.\ observed} & \textbf{Measurable impact} & \textbf{Affected agents} \\
\hline
D1 & Environment/agent RNG boundary; shared stream consumed at agent-dependent rates & Expected: same environment trace for a matched seed. Observed: episode-level station-failure count profile diverges although demand matches. & Demand (\texttt{n\_ev\_launched}) matches in 42/42 pairs; 42/42 pairs have different 15-episode profiles; cumulative totals differ in 34/42 (81.0\%; mean abs.\ diff.\ 3.90, max 16); performance effect not isolated. & Any agent sharing the RNG stream \\
D2 & Reward ablation flag; clip stays active when the fix flag is off & Expected: off path reproduces the pre-fix reward. Observed: off path inherits clipping. & Code-path contamination demonstrated; KPI magnitude not isolated. & Agents using the shared clipped reward path \\
D3 & State-to-reward interface; already-normalised distance divided again & Expected: detour term has declared non-trivial scale. Observed: term collapses to $\sim$$10^{-4}$ of intended scale. & Reward-term magnitude quantified; KPI effect not isolated. & DQN, DDQN \\
D4 & Event counter/replay insertion; persistent condition fires once per step & Expected: one rejection/penalty per vehicle event. Observed: repeated count/transition over consecutive steps. & At least 125.4$\times$ inflation in one DQN configuration (full detail); 93.8--103.1$\times$ mean (Rabat), 8.9--12.5$\times$ mean (Tangier), recalculated across the four non-random agents from the pre-audit archive. & Quantified for the four non-random agents (Rabat/Tangier means); full detail for DQN only \\
D5 & State builder/network input; four documented power features omitted & Expected: 17D documented state reaches the network. Observed: implemented network receives 13D. & Architectural mismatch established; on a fixed decision corpus (7 seeds, 499 states), greedy actions disagree at 16.6\% (DQN)/16.3\% (DDQN), descriptive; no KPI effect distinguishable from zero after Holm correction ($n=7$). & DQN, DDQN \\
\hline
\end{tabularx}
\end{table}

\section{Results and Discussion}
\label{sec:results}

This section addresses \textbf{RQ1} (Sec.~\ref{sec:research-questions}) by presenting empirical evidence on confirmed defect mechanisms, their
measured magnitudes, and the affected agent architectures. In turn, \textbf{RQ2} is answered through the diagnostic checklist
(Table~\ref{tab:checklist}) and its prospective validation (Sec.~\ref{sec:prospective}--\ref{sec:boundary-defects}).
\subsection{Quantified impact of D4}

In one representative configuration --- DQN agent, Rabat evening-peak
scenario, single seed, 55 episodes, pre-fix production data --- defect D4
drove the mean recorded rejection count to 9,402 against 75 vehicles
launched per episode: a 125.4x inflation, with observed counts ranging
from 7,888 to 10,712.
Why a single seed suffices here, deliberately: what is being established
is a deterministic counting defect, a counter incremented once per
simulation step rather than once per event, not a statistical estimate of
performance --- the multi-seed requirements discussed in \cite{b6} apply
to the latter, not the former. The factor is also a conservative lower
bound, since the denominator counts all launched vehicles while only
vehicles that actually requested charging could be rejected; the true
per-event inflation therefore exceeds the value reported.
Extending the same recalculation across the four non-random agents and
both cities (Table~\ref{tab:d1d5-mechanism}, row D4) yields a
substantially lower mean inflation at Tangier (8.9--12.5$\times$) than at
Rabat (93.8--103.1$\times$). Both remain a severe, qualitatively
identical counting defect; what changes is that the factor is not
scenario-invariant, and we stop short of extrapolating the
single-configuration value past what this multi-city, multi-agent
recalculation shows. Beyond corrupting the reported KPI itself, the
defect floods the replay buffer with near-duplicate terminal transitions
during congestion --- plausibly biasing policy learning away from the
true congestion signal.

\subsection{Case study: partial fixes can still mislead ablations}
\label{sec:case-study}

The defects described above were identified in a later audit of this
benchmark, separate from an earlier round of reward-shaping refinements
applied to the same codebase. That earlier ablation --- comparing three
configurations of the DQN agent's reward function, A0 (an earlier
queue-wait redefinition only), A1 (A0 plus a rejection penalty), and A2
(A1 plus a quadratic congestion penalty, a raised congestion weight, and
a longer discount horizon) --- was originally reported, on its own
performance-oriented terms, as part of our companion deployment
study~\cite{b15}. We revisit it here under a different lens: not to
report the reward fix's performance, but to illustrate, with the
benefit of the later D1--D5 audit, a related and independent
reproducibility risk using a fix set unrelated to D1--D5 --- that a
seemingly clean ablation study can still mislead. The ablation spans
six (city, traffic-period) scenarios, Wilcoxon signed-rank test,
$n=7$ seeds, comparing queue-wait time (QW, the metric the A2 changes
specifically targeted) against three secondary KPIs (completion rate CR,
energy delivered ED, vehicles served) between configurations A1 and A2.

Start with the primary KPI, QW. No scenario reached statistical
significance: $p$ ranges 0.109--0.938 across the six scenarios, not
statistically distinguishable from zero at the available seven-seed
resolution. Absence of significance is not evidence of absence, though.
Seven matched pairs give exact Wilcoxon inference little room to move,
power stays limited, and D1 adds environment variation inside nominal
pairs (see below); given all that, we read the corrected non-significance
as insufficient evidence of an A2 effect, nothing stronger.

\begin{figure}[!t]
\centering
\includegraphics[width=0.88\linewidth]{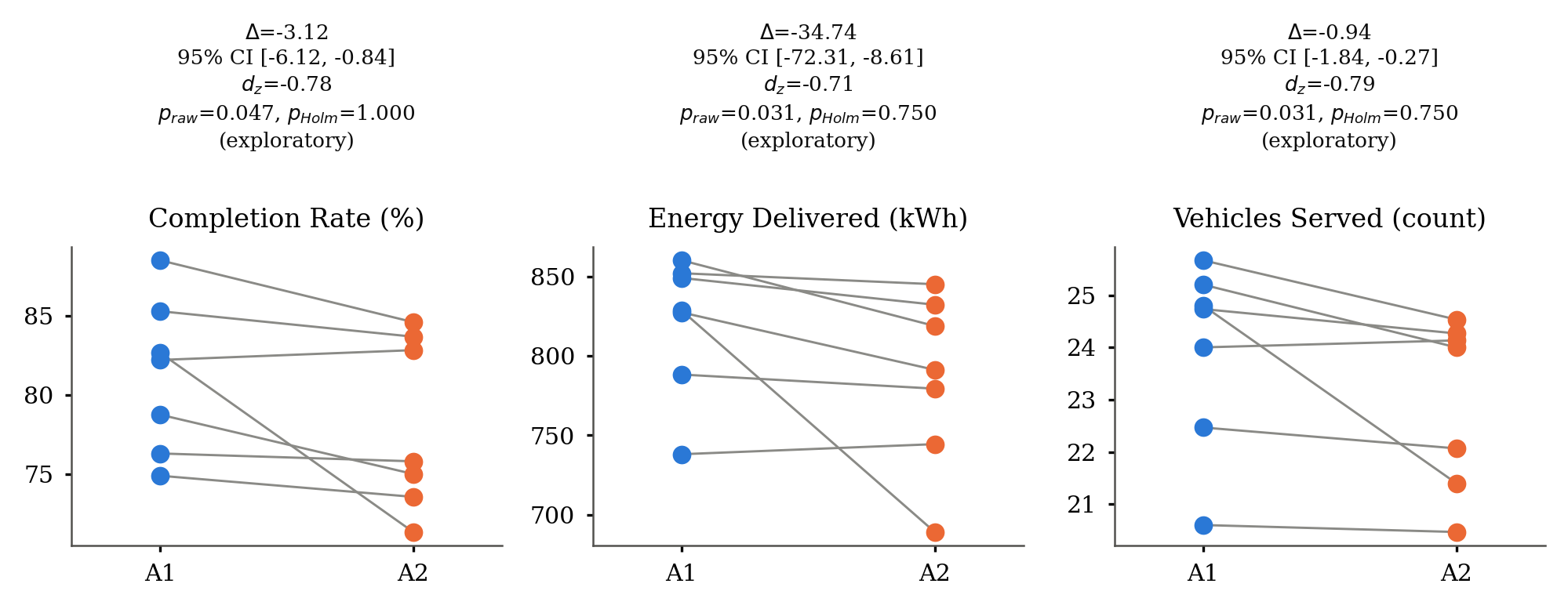}
\caption{Seed-paired secondary-KPI differences between A1 and A2 at
Tangier morning ($n=7$). Lines connect each seed's A1 and A2 values;
$\Delta$ is the mean paired difference with a percentile-bootstrap 95\% CI
(RNG seed 20260817), and $d_z$ is paired Cohen's $d$. The differences are
exploratory: none survives Holm--Bonferroni correction across the
24-comparison family (Table S2).}
\label{fig2}
\end{figure}

Figure~\ref{fig2} tells a different-looking story on three secondary
KPIs at Tangier morning, and every one of them moved the wrong way:
completion rate fell from 81.2\% to 78.1\% (unadjusted $p=0.0469$),
energy delivered from 820.5 to 785.8~kWh ($p=0.0312$), vehicles served
from 23.9 to 23.0 ($p=0.0312$). Exploratory, unadjusted signals --- that
is how we report them, not as evidence that A2 degraded performance.
And here correction genuinely decides the outcome rather than trimming
at the margins: across a family of 24 tests the first Holm threshold is
$0.05/24 = 0.00208$, and even the smallest unadjusted $p$-value on
record, $0.0312$, sits well above it. (Table S2 lists all 24 comparisons
with both their unadjusted and Holm-adjusted $p$-values, so this is
checkable directly.) One might reasonably ask whether the four-KPI
case-study family --- the narrower, more permissive choice, since a
smaller family softens the Holm correction --- was picked to produce this
outcome. It was not. Widen the family to every KPI in the underlying
synthesis table instead ($m=36$, first Holm threshold $0.05/36=0.00139$)
and the smallest adjusted $p$-value still lands at $1.000$, with zero
comparisons surviving. However the family is drawn, the conclusion does
not move.

Every $p$-value quoted above was recomputed independently from the
released seed-level measurements, via the distributed script, as an
exact conditional enumeration over all $2^{7}$ sign assignments: all 24
declared-family comparisons reproduce the values reported. One
comparison sits outside that family --- the Rabat-evening first-step
assignment rate --- and it differs
($0.9375$ versus $0.8125$); we report it for transparency, though it
does not enter the declared inference.

The re-analysis does not establish degradation; it illustrates how
directional secondary signals can mislead when multiplicity and later
code-audit findings are ignored, because a defect discovered after an
ablation can invalidate its premise without retroactive checking.

D1 can be seen in these runs: the A1 and A2 campaigns used the
shared-generator code before D1 was found. Across 42 matched pairs (6
scenarios $\times$ 7 seeds), vehicle demand matches in 42/42 pairs, but
every 15-episode station-failure count profile differs in at least one
episode; cumulative totals differ in 34/42 (81.0\%; mean absolute
difference 3.90, maximum 16). This is an episode-level count profile, not
a timestamped event trace: the archived CSVs contain neither event timing
nor station identity, so no finer-grained divergence metric is computable.
The independently seeded demand component matched; station failures sharing
the agents' generator did not, degrading the seed pairing and lowering test
power.

\subsection{D1 revisited: does the RNG fix change the ablation conclusion?}
\label{sec:a7-validation}

Sec.~\ref{sec:case-study} reports that the historical A1--A2 ablation's
seed pairing was already degraded by D1, which lowers the signed-rank
test's power without establishing what a corrected pairing would have
concluded. Per reviewer feedback on the 2026-08-21 decision, we quantify
this directly rather than leaving it as a plausible mechanism: a
dedicated 56-run validation campaign (2 scenarios [Rabat evening, Tangier
morning] $\times$ 2 reward configurations [A1, A2] $\times$ 2 RNG regimes
[\emph{independent} (the corrected D1 fix) and \emph{shared\_compat}, a
controlled, fail-closed reintroduction of the pre-fix shared stream
gated behind an explicit environment flag and a locked experiment
manifest] $\times$ 7 seeds; 40 training plus 15 evaluation episodes
sharing one checkpoint per run) isolates the RNG regime as the only
varying factor, everything else --- code, reward configuration,
scenario, seed --- held fixed.

Primary estimand: for each (scenario, KPI) cell and each seed, the
interaction contrast $I = [A2-A1]_{\text{shared\_compat}} -
[A2-A1]_{\text{independent}}$. This is not a repeat of the historical
A1--A2 test; it tests whether the D1 fix changes the A1--A2 effect that
test measured. Across the resulting family of 8 exact Wilcoxon
signed-rank tests ($n=7$ seeds; 2 scenarios $\times$ 4 KPIs: queue-wait
time, completion rate, energy delivered, vehicles served), no cell
survives Holm--Bonferroni correction (smallest $p_{\text{raw}} = 0.2188$,
all $p_{\text{Holm}} = 1.0$; paired effect sizes small to moderate,
$|d_z| \in [0.02, 0.55]$; 10,000-resample percentile bootstrap CI, RNG
seed 20260817). This is evidence of robustness at the available
resolution, not proof of exact equivalence: the D1 fix does not
detectably alter the non-significant A1--A2 conclusion already reported
in Sec.~\ref{sec:case-study}.

Secondary analysis: using event-level logging enabled specifically for
this campaign, we compare the station-failure/station-recovery trace
between the \emph{independent} and \emph{shared\_compat} arms for each
of the 28 matched (scenario, reward configuration, seed) triples. None
of the 28 pairs produce an identical train-phase event sequence (0/28);
in the earliest case examined, the two arms' first station failure
already disagrees within the opening minutes of episode 1. This is
direct, event-level confirmation --- unavailable from the historical
archive, whose episode-level counters cannot resolve event timing
(Sec.~\ref{sec:case-study}) --- that D1 is a demonstrated mechanism
rather than a plausible one: sharing the RNG stream measurably
desynchronises the environment's Markov process from a matched seed
alone.

Together, the two analyses answer the reviewer's request precisely: D1
demonstrably alters the environment trajectory (secondary analysis),
while the A1--A2 ablation conclusion it could have biased is not
measurably changed by correcting it (primary analysis). We report both
rather than only the reassuring one, because a null interaction result
at $n=7$ seeds is evidence of no detected effect, not evidence of no
effect, and does not retroactively certify every comparison run under
the historical shared-RNG protocol outside this specific 56-run
campaign.

\subsection{Prospective checklist detections: D6--D7}
\label{sec:prospective}

The checklist was subsequently applied prospectively, live, during a
pilot regeneration of the same benchmark under a corrected reward
pipeline. Two further silent defects distinct
from D1--D5 were exposed directly by checklist checks during that cycle:

\begin{itemize}[leftmargin=*,itemsep=1pt,topsep=2pt]
\item \textbf{D6} --- an action mask built as a list of feasible-action
  indices instead of a binary vector, crashing the masked-target
  computation on the first full station (checks 2, 6).
\item \textbf{D7} --- a vehicle-release guard deferring
  \texttt{resume()} whenever a vehicle was still inside the
  charging-station interval, true of every vehicle that had just
  finished charging: physical occupancy pinned at capacity across 585
  consecutive episode-rows, zero variance, caught only by comparing the
  ``off'' condition against direct simulator instrumentation rather than
  the agent's own bookkeeping (check 7).
\end{itemize}

Both were fixed and locked in with regression tests before the next
campaign (check 8). We state the evidential value of this cycle precisely,
because it is the only prospective evidence available: D1--D5 cannot count,
the checklist having been derived from them. The prospective cycle directly
exercised checklist checks 2, 6 and 7 through D6--D7.

\subsection{Boundary defects found outside the checklist: D8--D10}
\label{sec:boundary-defects}

D8--D10 are not validation successes of the eight-item checklist. They
were discovered by broader engineering inspection during the same
regeneration cycle and are reported to delimit the checklist's current
scope and motivate candidate future checks, not to demonstrate its
coverage.

\begin{itemize}[leftmargin=*,itemsep=1pt,topsep=2pt]
\item \textbf{D8} --- a multiprocess pool silently re-importing an
  unfiltered scope configuration under \texttt{spawn}, so a manifest
  declaring one traffic period ran all three.
\item \textbf{D9} --- an unconditional \texttt{--append} flag producing
  a headerless results file whenever an evaluation-only run skipped the
  training phase.
\item \textbf{D10} --- an unguarded stdout rewrap that only manifested
  under a test harness, silently corrupting its output capture.
\end{itemize}

All three were fixed and locked in with regression tests before the next
campaign (check 8). To avoid
retroactive validation, two candidate checks follow from them but are
presented as future extensions and are not credited with detecting
D8--D10: \emph{process-boundary configuration parity} ---
every spawned worker must receive and log the resolved canonical
configuration, and the expected task cardinality must be checked; and
\emph{execution-interface and artefact contracts} --- every execution mode
must be exercised end to end, with the CSV schema, header, parseability,
stdout/stderr separation and expected artefacts asserted. That five
defects of the same silent, comparison-biasing character appeared in a
single regeneration cycle, months after D1--D5 were fixed, supports
treating this as a recurring hazard class rather than a closed, one-time
list.

\subsection{Diagnostic checklist}

We distil the above into eight checks, ordered roughly by ease of
application; none requires re-running a full experimental campaign. The
format is deliberate: Pineau et al. \cite{b20}, reporting on the NeurIPS
2019 reproducibility programme, showed the value of short, actionable
reproducibility checklists adopted at conference scale. Ours is narrower
and complementary --- it audits the environment and reward implementation
rather than submission reporting --- operationalising, for a single
pipeline, the empirical-design principle Patterson et al. \cite{b21}
argue should govern RL experiments: the apparatus itself must be
validated before any comparison it produces can be interpreted. At least
four checks restate known practice: RNG separation (check~1) restates
Patterson et al.'s~\cite{b21} env--agent seed guidance; unit consistency
at reward boundaries (check~3) and cross-agent objective parity
(check~6) are the code-level scrutiny Engstrom et al.~\cite{b16} showed
matters more than headline algorithmic choices; re-verification after
each fix (check~8) is ordinary regression-testing hygiene. We claim no
antecedent for checks 2, 4, 5, 7 individually; what is not already
established practice, to our knowledge, is auditing the three
RNG/reward/state interfaces jointly, as one ordered pass, on a
SUMO-based EV-charging pipeline specifically --- of the ten works
compared in Sec.~\ref{sec:related-work}, only Patterson et
al.~\cite{b21} reports all three jointly, and not within a traffic- or
EV-charging-simulation pipeline, a claim
bounded to that comparison, not the wider literature. Each check also targets a specific audit stage:
checks 1 and 5 are pre-training assertions, checks 2, 3 and 7 are
reward-function unit tests independent of training, check 6 is a
fixed-transition sanity check across agents, and checks 4 and 8 apply
during and after a campaign respectively. Table~\ref{tab:checklist}
states each check as an executable assertion rather than a verb without
an oracle, with checks 1--5, 7, and 8 giving a concrete pass/fail
criterion; check 6 has no numeric pass threshold yet for ``comparable
magnitude'' in the audited pipeline and is marked \emph{oracle not yet
specified} rather than machine-checkable, so as not to present an
unenforced criterion as executable --- a project adopting it must first
fix its own tolerance (e.g.\ relative to the largest active reward
term). An evidence artefact marked \emph{not yet included} is a check we
run in the audited codebase's own test suite but do not ship in the
reproducibility archive; it is reported as
such rather than presented as an already-distributed test.

\begin{table}[!t]
\centering
\fontsize{7.5pt}{9.5pt}\selectfont
\caption{Eight-item diagnostic checklist as executable assertions. No full
campaign rerun is required.}
\label{tab:checklist}
\begin{tabularx}{\textwidth}{L{0.4cm}L{2.0cm}YL{1.0cm}L{5.0cm}}
\hline
\textbf{\#} & \textbf{Invariant} & \textbf{Executable assertion / pass criterion / evidence} & \textbf{Mapped defect(s)} & \textbf{Corrective action} \\
\hline
1 & RNG separation & Two agents, same seed $\Rightarrow$ identical environment-event hash. Evidence: covered by the project's regression suite; the 56-run validation campaign (Sec.~\ref{sec:a7-validation}) confirms directly that violating this invariant desynchronises the trace (0/28 identical sequences between regimes). & D1 & Dedicated, independently-seeded environment RNG; traces identical across agents for every matched seed \\
2 & Ablation flag coverage & Every downstream operation of a conditional fix gated by the same flag as the primary computation; ``off'' path bit-identical to a frozen pre-fix fixture. Evidence: not yet included. & D2, D6, D7 & Gate clipping and every side effect under the same flag; off path bit-identical to the pre-fix fixture \\
3 & Unit consistency at reward boundaries & Known raw input $\Rightarrow$ reward term falls within a declared numeric interval. Evidence: not yet included. & D3 & Separate raw reward-facing and normalised state-facing values; term magnitude in the declared interval for representative inputs \\
4 & Per-event vs.\ per-step counters & One persistent rejected/failed entity across multiple timesteps $\Rightarrow$ counter increments exactly once. Evidence: not yet included. & D4 & Per-episode \texttt{already\_counted} guard; exactly one counter increment/penalty per entity event \\
5 & Documented vs.\ implemented state/action space & \texttt{len(state) == network\_input\_dim == documented\_dim} for every agent and state mode. Evidence: locked by regression tests (Sec.~\ref{sec:taxonomy-intro}, D5). & D5 & Single derived dimension constant; state length, network input, and documented dimension are equal \\
6 & Cross-agent objective parity & On one fixed transition, every agent's realised reward has the same active terms; magnitude comparability has \emph{no numeric threshold fixed yet} (oracle not yet specified). Evidence: active-terms comparison alone caught D6 in the prospective cycle (Sec.~\ref{sec:prospective}), since a crash needs no numeric threshold; the magnitude criterion is not yet included. & D3, D6 & See check~3 for D3; D6 required no numeric fix, only the active-terms comparison itself \\
7 & Ablation-condition purity & ``Off'' configuration hash/trace identical to a known pre-fix baseline on a declared fixture. Evidence: not yet included as an assertion; D7 was caught in the prospective cycle (Sec.~\ref{sec:prospective}) via direct simulator instrumentation. & D2, D7 & See check~2 for D2; D7 fixed via direct simulator instrumentation, no generic fixture yet \\
8 & Re-verify after each fix & Each individual defect fix ships with its own smoke test before being folded into a full campaign. Evidence: exercised for D6--D10 (Sec.~\ref{sec:prospective}). & D6--D10 & Individual fix plus smoke test per defect (Fig.~S1); no single generic corrective action \\
\hline
\end{tabularx}
\end{table}

This checklist runs as a loop, not a one-pass list: the eight checks
execute in order, and a failure at any step triggers a minimal fix
followed by a smoke test, after which the workflow restarts from check~1
--- since a fix can invalidate a check that had already passed --- until
all eight pass in sequence. This mirrors how the defects reported above
were actually found and corrected in the audited codebase. The full
diagram, as a UML activity workflow looping over the check index, is
provided as Figure S1.

\subsection{Limitations}
\label{sec:limitations}

The checklist targets silent defects that bias agent or configuration
comparisons, not general software-engineering hygiene; it excludes training
instability, architecture search, and simulator artefacts outside the RL
pipeline. Training-environment reproducibility is partial: the supplementary
parameter table reports the SUMO version, reward function, and principal hyperparameters,
but historical NumPy, TraCI, and optional MQTT versions were never pinned
and cannot be recovered without fabricating a record; this gap is reported
explicitly. The taxonomy is project-specific and non-exhaustive. D4's
125.4x rejection-counter inflation
(the representative calculation; Table~\ref{tab:d1d5-mechanism}) recurs across agents and
cities but remains scenario-dependent: it does not establish a per-episode
distribution or a causal policy-level KPI effect, and we do not extrapolate
beyond the two reported summary statistics. The 56-run D1 interaction-contrast
campaign (Sec.~\ref{sec:a7-validation}) re-evaluated one A1--A2 ablation
across two scenarios, seven seeds, and two RNG regimes; it does not certify
every historical shared-RNG comparison outside that campaign, and its null
result at this seed count cannot exclude smaller effects. Independent-codebase
validation and a causal, multi-agent D4 extension remain future work
(Sec.~\ref{sec:conclusion}). The checklist is a low-cost first pass (each
check is executable in well under a day), not a substitute for independent
re-implementation or replication, nor for the broader empirical-design
guidance of Jordan et al.~\cite{b19,b32}.

\section{Conclusion}
\label{sec:conclusion}

We documented five silent defect patterns affecting reward functions,
environment stochasticity, and state encoding in RL-traffic-simulation
benchmarks, quantified their impact on a real five-agent EV-charging
benchmark --- including an at-least-125.4$\times$ rejection-counter
inflation in the representative Rabat-evening configuration
(8.9--103.1$\times$ across agents and cities, D4), a reward term neutralised to about $10^{-4}$ of its
intended scale (D3), and a degraded, shared-RNG seed pairing affecting
the very comparisons a companion ablation relied on (D1) --- and
re-analysed that historical reward ablation, in which no effect survived
family-wise correction --- evidence that even careful,
multi-seed benchmarks can yield directional but non-robust signals when
multiplicity control and a subsequent code audit are absent. This
five-defect taxonomy and the checklist derived from it are a starting
inventory from a single audited pipeline, not an exhaustive or
universally applicable classification; whether they recur, and at what
rate, in other RL-traffic-simulation codebases remains to be established
by future work; the same non-exhaustive, single-pipeline status
applies to the eight-item checklist we offer as a practical, low-cost
starting point for researchers building similar pipelines. We report
partial prospective evidence for its value (Sec.~\ref{sec:prospective}):
applied live to the
same codebase during a subsequent pilot regeneration, checks~2, 6 and~7
exposed two further silent defects, while three additional process and
artefact defects surfaced through broader engineering inspection and
delimit the checklist's present coverage. All five were fixed and locked
in with regression tests before being folded into the next campaign.
Future work will extend the checklist to these uncovered classes and
apply it to independent codebases, to assess whether the D1--D10 patterns,
and their apparent recurrence rate, generalise beyond this one project's
benchmark. For researchers and developers building similar
EV-charging or traffic-simulation RL pipelines, the practical
implication is procedural: run the eight checks before trusting any KPI
comparison the pipeline produces, and re-run the checks a fix touches
after every subsequent change to the pipeline (Fig.~S1),
since each is inexpensive relative to a full campaign and D1--D10 show
that a plausible-looking KPI
table is not, by itself, evidence that no such defect is present.

\section*{ACKNOWLEDGMENTS}
The authors acknowledge support from the Laboratory of Research in Computer
Science and Telecommunications (LRIT), Mohammed V University, Rabat, and the
Laboratory of Probability, Statistics and Intelligent Systems (LPAIS), Sidi
Mohamed Ben Abdellah University, Fez, Morocco.

\section*{FUNDING INFORMATION}
This research received no specific grant from funding agencies in the public, commercial, or not-for-profit sectors.

\section*{AUTHOR CONTRIBUTIONS STATEMENT}
\begin{table}[H]
	\centering
	\selectfont
	\begin{tabular}{lcccccccccccccc}
		\hline
		\textbf{Name of Author}\quad\quad\quad\quad\quad\quad & \cellcolor[HTML]{D8F0F8}\textbf{C} & \cellcolor[HTML]{D8F0F8}\textbf{M} & \textbf{So} & \textbf{Va} & \cellcolor[HTML]{D8F0F8}\textbf{Fo} & \cellcolor[HTML]{D8F0F8}\textbf{I} & \textbf{R} & \textbf{D} & \cellcolor[HTML]{FDE9D9}\textbf{O} & \cellcolor[HTML]{FDE9D9}\textbf{E} & \textbf{Vi} & \textbf{Su} & \textbf{P} & \textbf{Fu} \\
		\hline
		Nour-Eddine Moumni & \cellcolor[HTML]{D8F0F8}\checkmark & \cellcolor[HTML]{D8F0F8}\checkmark & \checkmark & \checkmark & \cellcolor[HTML]{D8F0F8}\checkmark & \cellcolor[HTML]{D8F0F8}\checkmark &  & \checkmark & \cellcolor[HTML]{FDE9D9}\checkmark & \cellcolor[HTML]{FDE9D9}\checkmark &  &  &  &  \\
		Rachid Alaoui & \cellcolor[HTML]{D8F0F8} & \cellcolor[HTML]{D8F0F8} &  &  & \cellcolor[HTML]{D8F0F8} & \cellcolor[HTML]{D8F0F8} &  &  & \cellcolor[HTML]{FDE9D9} & \cellcolor[HTML]{FDE9D9}\checkmark &  & \checkmark & \checkmark &  \\
		Driss Kiouach & \cellcolor[HTML]{D8F0F8} & \cellcolor[HTML]{D8F0F8} &  &  & \cellcolor[HTML]{D8F0F8} & \cellcolor[HTML]{D8F0F8} &  &  & \cellcolor[HTML]{FDE9D9} & \cellcolor[HTML]{FDE9D9}\checkmark &  & \checkmark &  &  \\
		\hline
		\multicolumn{15}{c}{
				\begin{tabular}{llllll}
				&&&&&\\
				C &: 	\textbf{C}\fontsize{8}{10}\selectfont onceptualization \quad\quad\quad\quad\quad\quad& I &: 	\textbf{I}\fontsize{8}{10}\selectfont nvestigation & Vi&: 	\textbf{Vi}\fontsize{8}{10}\selectfont sualization  \\
				M &: 	\textbf{M}\fontsize{8}{10}\selectfont ethodology & R &: 	\textbf{R}\fontsize{8}{10}\selectfont esources & Su&: 	\textbf{Su}\fontsize{8}{10}\selectfont pervision  \\
				So&: 	\textbf{So}\fontsize{8}{10}\selectfont ftware & D &:	\textbf{D}\fontsize{8}{10}\selectfont ata Curation & P &: 	\textbf{P}\fontsize{8}{10}\selectfont roject Administration  \\
				Va&: 	\textbf{Va}\fontsize{8}{10}\selectfont lidation & O &:	\fontsize{8}{10}\selectfont Writing - \fontsize{10}{12}\selectfont \textbf{O}\fontsize{8}{10}\selectfont riginal Draft & Fu&: 	\textbf{Fu}\fontsize{8}{10}\selectfont nding Acquisition \\
				Fo&: 	\textbf{Fo}\fontsize{8}{10}\selectfont rmal Analysis & E &:	\fontsize{8}{10}\selectfont Writing - Review \& \fontsize{10}{12}\selectfont\textbf{E}\fontsize{8}{10}\selectfont diting \quad\quad\quad\quad\quad\quad&  \\
			\end{tabular}
		}
	\end{tabular}
\end{table}

\section*{CONFLICT OF INTEREST STATEMENT}
The authors declare that they have no known competing financial interests or personal relationships that could have appeared to influence the work reported in this paper.

\section*{DATA AVAILABILITY}
A supplementary reproducibility archive --- the raw campaign data,
seed-level measurements, figures, SHA-256 manifests, and supplementary
Tables S1--S3, including the frozen D5 corpus and the D1
validation-campaign data --- is permanently archived on Zenodo, DOI:
\href{https://doi.org/10.5281/zenodo.22166118}{10.5281/zenodo.22166118}.
The remainder of the archive is available from the corresponding author
on request, and is planned for a later version of the same Zenodo
record.

\footnotesize

\bibliographystyle{IEEEtran}
\setlength{\bibsep}{0pt}
\bibliography{references_BEEI}

\section*{BIOGRAPHIES OF AUTHORS}
\normalsize
\vspace{-.7em}
\small
\begin{biography}[{\includegraphics[width=2.4cm,height=4cm,clip,keepaspectratio]{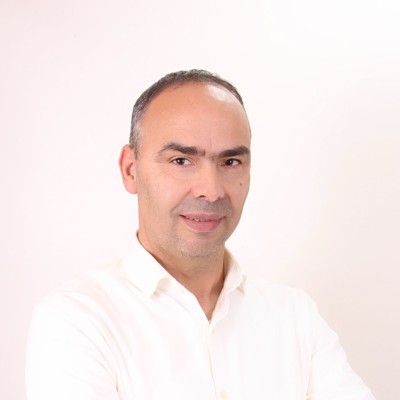}}]
\textbf{Nour-Eddine Moumni}
\href{https://orcid.org/0009-0008-5803-0604}{\includegraphics[width=0.017\textwidth]{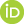}}
\href{https://www.scopus.com/authid/detail.uri?authorId=58847408500}{\includegraphics[width=0.017\textwidth]{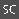}}
is a PhD candidate at the Laboratory of Research in Computer Science and
Telecommunications (LRIT), Mohammed V University, Rabat, Morocco. His
research addresses reinforcement learning for EV charging management in
smart cities --- reward-function design, multi-agent coordination, and
reproducibility of microscopic-traffic-simulation benchmarks --- and
develops an agent-based charging-assignment framework coupling SUMO with
IoEV communication on real Moroccan OpenStreetMap networks; validating a
simulation pipeline before trusting its comparisons motivated the
diagnostic checklist presented here. Current interests include
prospectively specified evaluation protocols for applied RL.
Contact: noureddine\_moumni@um5.ac.ma.
\end{biography}
\vspace{-.7em}

\begin{biography}[{\includegraphics[width=2.4cm,height=4cm,clip,keepaspectratio]{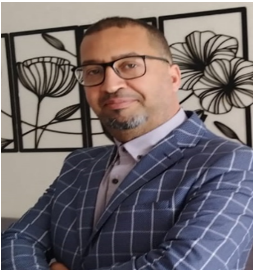}}]
\textbf{Rachid Alaoui}
\href{https://orcid.org/0000-0001-9948-329X}{\includegraphics[width=0.017\textwidth]{orcid.png}}
\href{https://www.scopus.com/authid/detail.uri?authorId=57214771728}{\includegraphics[width=0.017\textwidth]{scopus.png}}
is a Full Professor of Higher Education at the Laboratory of Research in
Computer Science and Telecommunications (LRIT), Mohammed V University,
Rabat, and at the Higher School of Technology of Sal\'e (EST Sal\'e),
Morocco. His research spans biometric systems, intelligent systems, and
smart electric vehicle charging: deep-learning multimodal biometric
identification (convolutional fusion of fingerprint, finger-vein, and face
images; score-level fusion for human recognition), smart coordination
systems minimising EV trip duration and peak charging demand, and
simulation platforms for optimising EV routes on Moroccan urban networks.
That experience in simulation-based mobility studies underpins his role in
the present work, where he supervised the audit and the framing of its
findings as a reusable checklist. He is indexed on Scopus with 25
documents, an h-index of 8, and more than 250 citations (August 2026).
Contact: rachid.alaoui@est.um5.ac.ma.
\end{biography}
\vspace{-.7em}

\begin{biography}[{\includegraphics[width=2.4cm,height=4cm,clip,keepaspectratio]{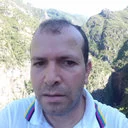}}]
\textbf{Driss Kiouach}
\href{https://orcid.org/0000-0002-1497-3028}{\includegraphics[width=0.017\textwidth]{orcid.png}}
\href{https://www.scopus.com/authid/detail.uri?authorId=26530532100}{\includegraphics[width=0.017\textwidth]{scopus.png}}
is a Full Professor at the Laboratory of Probability, Statistics and
Intelligent Systems (LPAIS), Faculty of Sciences Dhar El Mahraz, Sidi
Mohamed Ben Abdellah University, Fez, Morocco. His research lies in
stochastic processes, applied mathematics, and mathematical epidemiology,
with a sustained focus on the stability of nonlinear dynamical systems; he
has published widely on stochastic epidemic models, including global
stability analyses of SIRS models with generalised nonlinear incidence and
vaccination, threshold behaviour under vertical transmission, and the
influence of L\'evy noise on contagion dynamics. This background motivates
his contribution here: the paper's environment stochasticity, seed-paired
comparison, and family-wise error control rest on the same foundations,
and a stochastic model whose randomness is not controlled cannot support
the inference drawn from it. He has also co-authored smart coordination
and highway charging-plan management systems for electric vehicles on
Moroccan networks. He is indexed on Scopus with 43 documents, an h-index
of 14, and more than 880 citations (August 2026).
Contact: d.kiouach@uiz.ac.ma.
\end{biography}

\end{document}